\documentclass[journal]{IEEEtran}
\usepackage[utf8]{inputenc}
\usepackage[table,xcdraw]{xcolor}
\usepackage{marvosym,amsmath,amssymb,mathtools,tikz}
\usepackage{enumitem}

\usepackage{multirow}
\usepackage{graphicx}
\usepackage{subcaption}
\usepackage{siunitx,url}

\usepackage{newtxtext}

\providecommand{\LDP}{\text{LDP}}
\providecommand{\maj}{\text{maj}}
\providecommand{\tr}{\mathcal{TR}}
\providecommand{\ts}{\mathcal{TS}}
\providecommand{\DF}{\text{DF}}
\providecommand{\FtF}{\text{F2F}}
\providecommand{\FSW}{\text{FSW}}
\providecommand{\OR}{\text{OR}}

\providecommand{\dir}{\rightarrow}
\providecommand{\inv}{\leftarrow}
\providecommand{\bid}{\leftrightarrow}

\begin{document}

\title{Dynamic texture analysis for detecting \\ fake faces in video sequences}

\author{Mattia~Bonomi$^\dagger$, Cecilia~Pasquini
	and Giulia~Boato
	\thanks{$^\dagger$ Corresponding author. \newline
	M. Bonomi, C. Pasquini and G. Boato are with the Department of Information Engineering and Computer Science, University of Trento, Trento 38123, Italy (e-mail: mattia.bonomi@unitn.it, giulia.boato@unitn.it). C. Pasquini was with the Department of Computer Science, Universit\"{a}t Innsbruck, Innsbruck 6020, Austria (e-mail: cecilia.pasquini@uibk.ac.at).}
}

\maketitle


\begin{abstract}

The creation of manipulated multimedia content involving human characters has reached in the last years unprecedented realism, calling for automated techniques to expose synthetically generated faces in images and videos.

This work explores the analysis of spatio-temporal texture dynamics of the video signal, with the goal of characterizing and distinguishing real and fake sequences. We propose to build a binary decision on the joint analysis of multiple temporal segments and, in contrast to previous approaches, to exploit the textural dynamics of both the spatial and temporal dimensions. This is achieved through the use of Local Derivative Patterns on Three Orthogonal Planes (LDP-TOP), a compact feature representation known to be an important asset for the detection of face spoofing attacks.

Experimental analyses on state-of-the-art datasets of manipulated videos show the discriminative power of such descriptors in separating real and fake sequences, and also identifying the creation method used. Linear Support Vector Machines (SVMs) are used which, despite the lower complexity, yield comparable performance to previously proposed deep models for fake content detection.

\end{abstract}

\section{Introduction}
\label{sec:intro}

Being able to ensure and verify the integrity of digital multimedia content is recognized as an essential challenge in our society. In the last decade, the field of multimedia forensics has worked towards developing increasingly effective technological safeguards to address these issues, with the goal of inferring information on the acquisition settings and digital history of the images and videos under investigation. 

In parallel, computer graphics and machine vision have achieved impressive advances in the very last years in the creation of highly realistic synthetic audio-video content. Convincing digital representations of human characters appearing almost indistinguishable from real people can now be obtained automatically through increasingly accessible tools. These technologies are progressing at a tremendous pace, and can be coupled with advances in the field of text-to-speech synthesis.
While offering exciting opportunities for entertainment and content creation purposes, it is clear that such technologies can have significant security implications in different application scenarios.
As a matter of fact, digital versions of human faces are constantly streamed through video chats, video conferencing services, media channels, and even used for authentication purposes in replacement of traditional schemes based on fingerprints or passwords.


Thus, the need for forensic techniques able to deal with this new powerful manipulations has become of primary importance, leading to huge efforts and initiatives in developing robust forensic detection methodologies and benchmarking them on common datasets \cite{ff,deepfake_challenge}. While the identification of computer-generated faces has been widely addressed in the last decade, the data produced by advanced and AI-based creation tools have raised renovated attention due to the higher level of hyper-realism \cite{Vincent2019}, as well as new and more complicated technical challenges. In response, a number of new detection approaches have been proposed, with special focus on still images depicting synthetic faces. However, the problem of detecting fake characters in video sequences has been faced only very recently, since the quality of AI-generated videos depicting faces achieved only in the last couple of years a good level of perceptual quality and realism.
Currently, video forensics approaches developed for this problem mostly apply detection techniques designed for still images to single frames of the video sequence, often relying on deep representations of the pixel domain. However, in doing so they do not exploit the temporal information provided by video sequences, which might contain useful statistical characterizations and contribute to the detection capabilities of an automatic detector.

The analysis of discriminative cues over time is tackled by a few previous works. One direction is to detect behavioural anomalies of the face dynamics, like the absence of physiological signals \cite{conotter2014}, inconsistent expression patterns \cite{farid2019}, irregular eye-blinking \cite{blink2018}.
While in principle these methods are robust to geometric degradations and easily interpretable, their effectiveness is highly dependent on the scene content, as it is based on few semantic cues that might not be available in all video sequences. Also, deep learning machinery (like recurrent neural networks \cite{sabir2019,guera2018,amerini2019}) has very recently been used to model short frame sequences, showing promising results at the price of low interpretability, a typical issue of deep learning based approaches. Moreover, deep learning based techniques present the classical drawback of requiring careful training on a large and diverse amount of data to achieve transferability of results and to avoid overfitting. 

In this work, we aim at exploiting both texture and temporal information of the video sequence, by tackling an intermediate approach that relies on hybrid descriptors operating in both spatial and time domain. This yields relatively small feature representations that can be learned through simpler classifiers, such as linear SVMs.
While such descriptors have been successfully used for video-based face spoofing detection \cite{as_survey}, to the best of our knowledge their effectiveness has never been explored in the context of manipulated faces detection, although the two problems present significant analogies.
Our approach employs so-called Local Derivative Patterns on Three Orthogonal Planes (LDP-TOP), a variant of local binary patterns that operates on three dimensions and proved to be particularly effective in face anti-spoofing. Moreover, we propose to perform the analysis of entire video sequences by combining the predictions computed on multiple temporal segments, which proves to bring a significant accuracy gain. 


The remainder of the paper is structured as follows: Section \ref{sec:soa} summarizes the state of research for the problem of manipulated video and image faces detection; in Section \ref{sec:method}, we illustrate how the feature descriptors are extracted from single videos, while Section \ref{sec:cl} describes the proposed classification framework. Experimental results are reported in Section \ref{sec:results} and conclusions are drawn in Section \ref{sec:conc}.


\begin{figure*}[t!]
	\centering
	\begin{tikzpicture}[>=stealth]
	\node {\includegraphics[width=\textwidth]{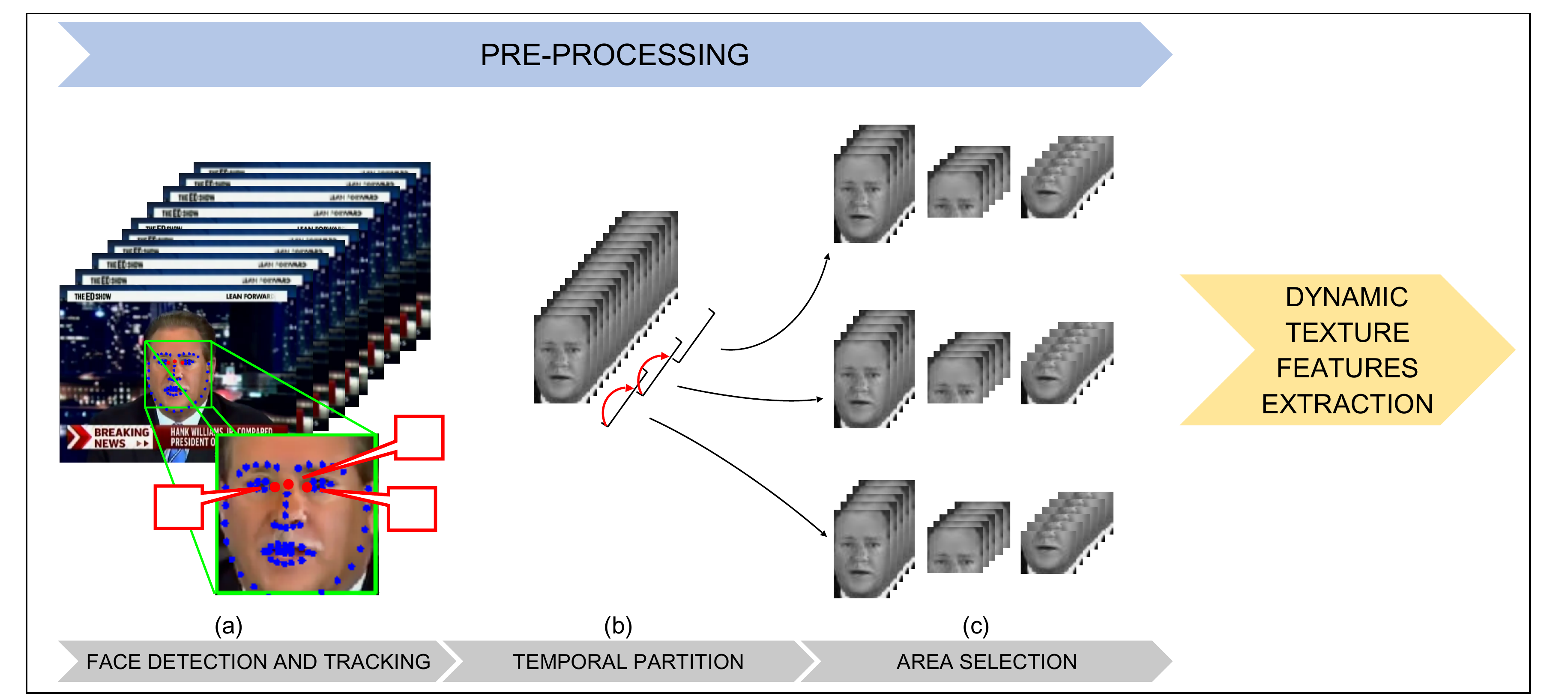}};
	
	\node at (-7.1,-1.85) {$\boldsymbol{l}$};
	\node at (-4.36,-1.86) {$\boldsymbol{r}$};	
	\node at (-4.26,-1.04) {$\boldsymbol{n}$};	
	
	\node at (-1.75,-0.75) {\scriptsize$d$};	
	\node at (-1.4,-0.4) {\scriptsize$d$};		
	\node at (-0.9,0.05) {\scriptsize$d$};			
	
	\node[red] at (-1.98,-0.65) {\scriptsize$s$};			
	\node[red] at (-1.54,-0.25) {\scriptsize$s$};			
	
	\node at (1.2,2.85) {\small$F$};			
	\node at (2.3,2.85) {\small$T$};			
	\node at (3.5,2.85) {\small$B$};

	\draw[->,gray] (0.55,1.15) -- node[below] {\scriptsize$W$} (1.25,1.15);
	\draw[->,gray] (0.55,1.15) -- node[left] {\scriptsize$H$} (0.55,2.5);
	\draw[->,gray] (1.25,1.15) -- node[right] {\scriptsize$H$} (1.55,1.5);		
	
	\end{tikzpicture}
	
	\caption{Workflow of the proposed pre-processing pipeline.}
	\label{fig:visualrepresentation}
\end{figure*}
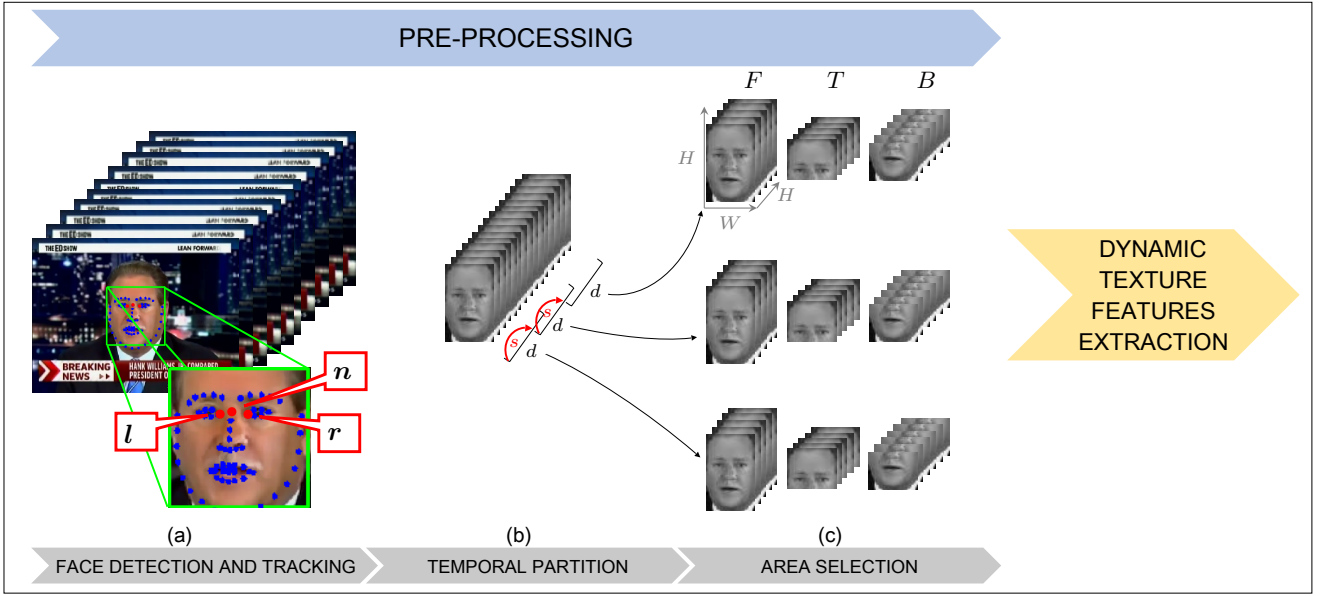

\section{Detection of manipulated faces in images and videos}
\label{sec:soa}

In order to position our work with respect to existing literature, this section briefly reviews the main classes of methods proposed for detecting manipulated and computer-generated (CG) faces in multimedia data.


\subsection{Methods based on statistical hand-crafted features}
Several methods proposed to distinguish real from manipulated multimedia content by exploiting statistical features capturing intrinsic properties of the media object.
Earlier works study specific traces that are present in real data due to operations at acquisition time \cite{Ng2005}, such as color filter array interpolation \cite{Gallagher2008}, or lens chromatic aberration \cite{Dirik2007}. 
Other approaches extract statistical features capturing the characteristics of the spatial texture \cite{Pan2009} \cite{Ke2013} and the coefficients distribution in transformed domains (e.g., wavelet) \cite{Lyu2005} \cite{Chen2009}, leading to supervised classification frameworks combining these cues \cite{Peng2016}.

More recently, detectors based on Fourier analysis coupled with conventional machine learning have been proposed also for modern AI-based manipulations \cite{fourier2019}. Such methods are applied to images only, thus they do not deal with the temporal evolution of video signals. As detailed in Section \ref{sec:method}, our work fills this gap by proposing a spatio-temporal texture description.

\subsection{Methods based on deep neural networks}

Deep neural networks are not only used for creation purposes but also as powerful tools for detecting fake content.

Several studies have been conducted on the use of deep networks to detect fake images generated by Generative Adversarial Networks (GANs) \cite{icarl2019,upsampling2019,marra2018}, and identify fingerprint specific GANs may leave \cite{attribution2019}.


A number of Convolutional Neural Networks (CNN) architecture have been proposed for the detection of manipulated faces videos, with the goal of characterizing artifacts arising when generating fake content. 
The authors in \cite{Rahmouni2017,Afchar2018} propose two shallow CNNs architectures exploiting mesoscopic features. In \cite{ff}, it is shown that deeper general-purpose networks like XceptionNet  in the same supervised scenario generally outperform shallow ones, as well as where re-adapted feature-based methods originating from steganalysis \cite{steg} and general-purpose image forensics \cite{bayar}. 

While these methods are applied individually on video frames, only few works operate along the temporal dimension. This is done in \cite{sabir2019} and \cite{guera2018} through the use of recurrent neural networks. In  in \cite{amerini2019}, a CNN is used to estimate and analyze the optical flow field across frames.


Finally, several deep-learning techniques have been recently proposed for other security applications, including the analysis of surveillance videos \cite{Xiao2019}, and the detection of suspect videos through usage of blockchain and smart contracts \cite{Hasan2019}. 

While the mentioned approach deliver good results in supervised scenarios, they typically tend to overfit the training set and suffer from performance decrease when dealing with unseen manipulations \cite{busch2018}. Additional strategies are then necessary to increase generalization capabilities, such as attention mechanisms \cite{stehouwer2019} or segmentation modules \cite{nguyen2019}. We refer the reader to \cite{verdoliva2020} \cite{nguyen2019deep} for thorough surveys of the literature on the topic. 


\subsection{Methods based on semantic cues}

As an alternative to hand-crafted or self-learned features, a number of methods aims at characterizing semantic features differentiating real and manipulated content. 
The work in \cite{matern2019} extracts typical artifacts appearing in GAN-generated images, such as non symmetrical colors and shape (in eyes and ears) or badly rendered details (e.g., blurry teeth areas).

Earlier studies on rendered faces  exploited geometric properties of the face in the spatial \cite{DangNguyen2012} and temporal domain \cite{dang2015TIFS}. Further properties like inconsistencies in facial landmark locations \cite{yangIH2019}, head pose \cite{yangICASSP2019}, and eye-blinking \cite{oculi2018} have also been exploited for exposing fakes.

By relying on video magnification techniques \cite{Wu_TG_2012}, the techniques developed in \cite{conotter2014} and \cite{bonomi2020} estimate the pulse rate of the depicted subject from temporal skin color variations, and show that this physiological signal is typically flat when the subject face is manipulated or computer generated. Similar ideas are explored in \cite{fakecatcher} and \cite{predheart}, where deep networks are used for this purpose.

Moreover, recent approaches \cite{farid2019} study and characterize soft traits specific individuals have in reproducing facial expressions and head movements, which are hardly reproducible in manipulated content.


\section{Extraction of spatio-temporal textural features}
\label{sec:method}

The methodology proposed in this work is composed of a preprocessing phase and a feature extraction phase. These two processes are described in the following subsections. Obtained feature representation will be learned in the classification phase described in the next section.

\begin{figure}[b!]
	\begin{tikzpicture}
	\node (1) at (0,0) {\includegraphics[width=0.23\textwidth]{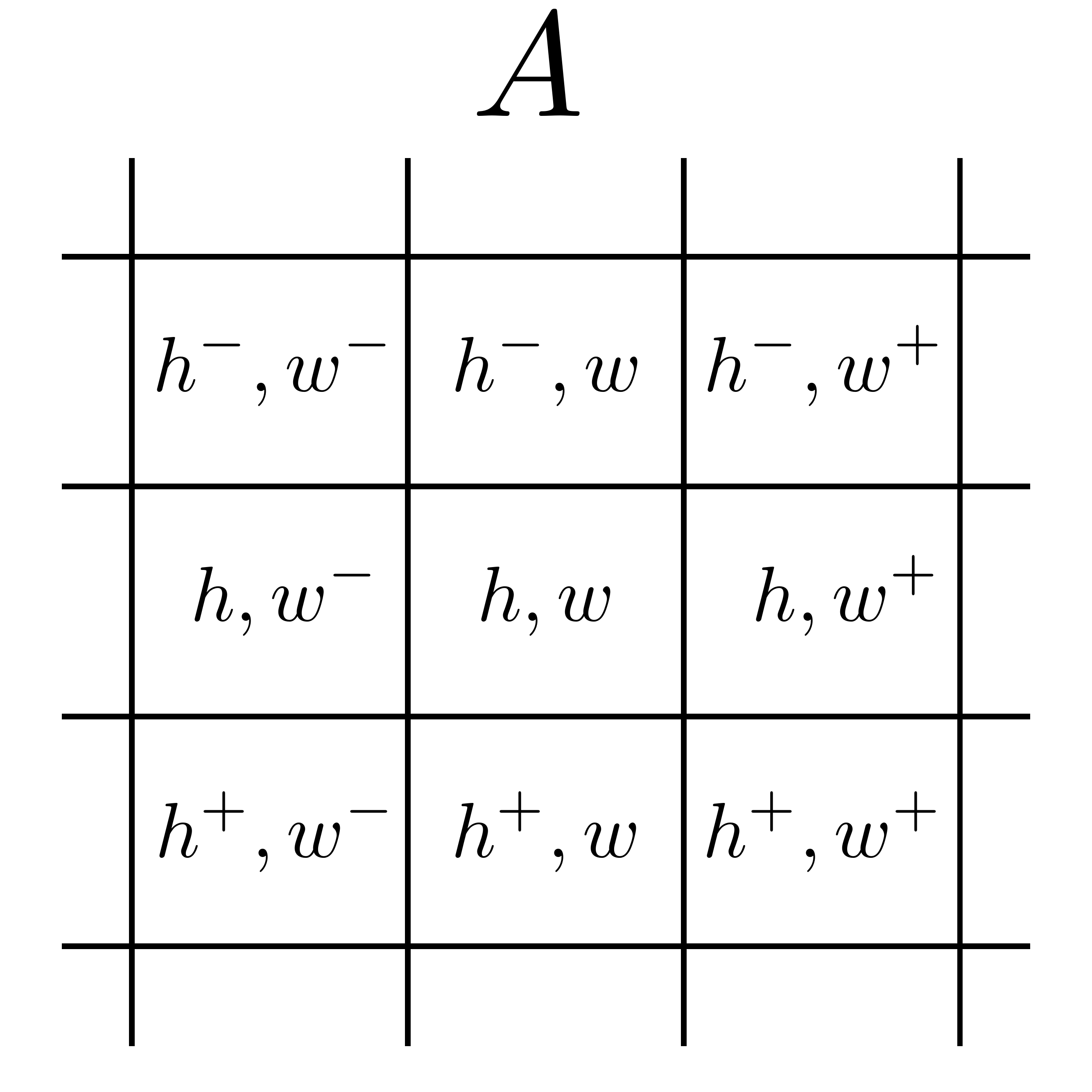}};
	\node[below=2.3cm] at (1) {(a)};
	
	\node[right=2.3cm] (2) at (1) {\includegraphics[width=0.25\textwidth]{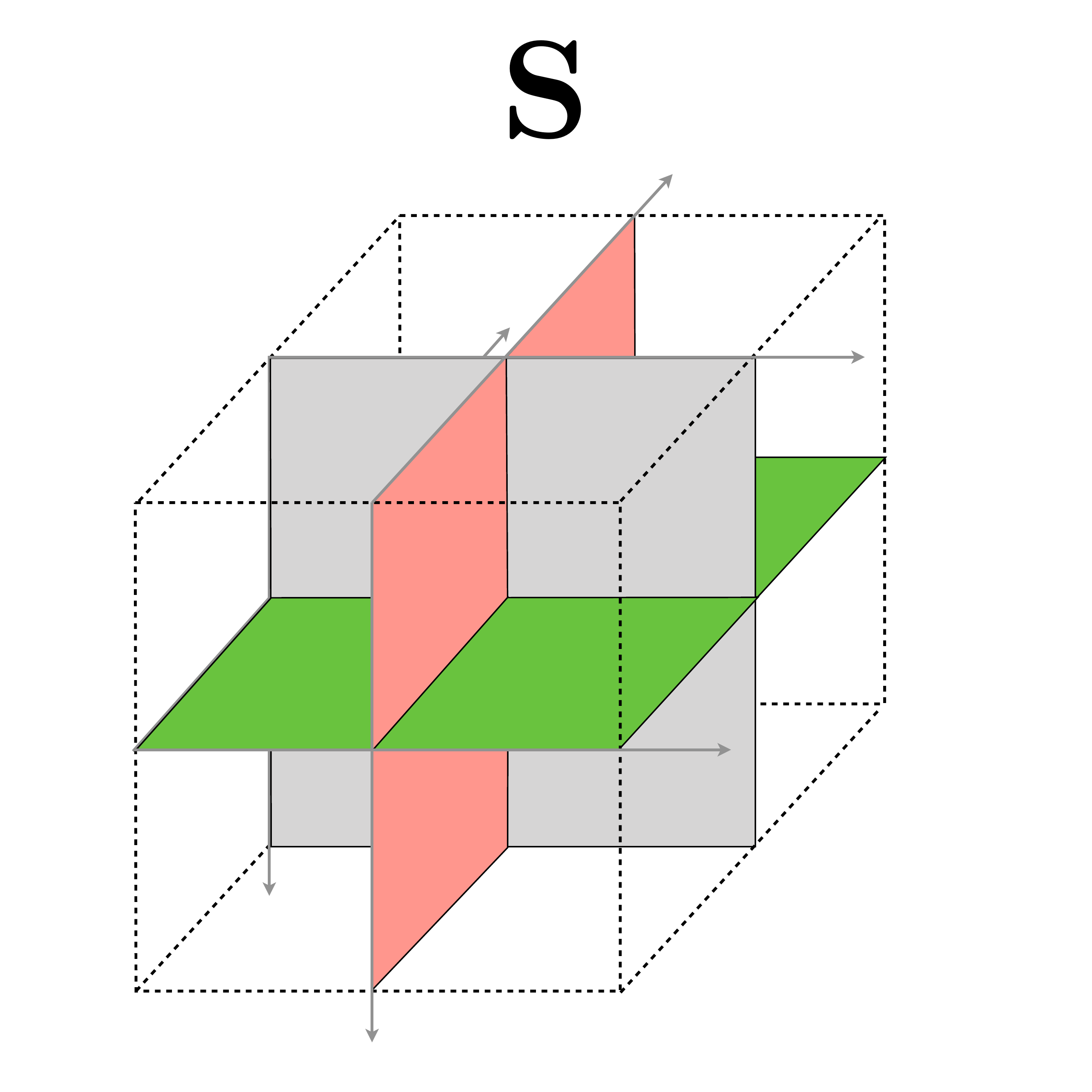}};
	\node[below=2.3cm] at (2) {(b)};
	
	\end{tikzpicture}
	
	\caption{Representations of the $3\times 3$ neighborhood and the three orthogonal planes used for the extraction of the LDP-TOP descriptors.}
	\label{fig:ldp}
\end{figure}

\subsection{Pre-processing}
\label{ssec:prep}

First, video patches are extracted and partitioned in multiple temporal sequences\footnote{Python 3.6.7 with the OpenCV2 4.1.0 libraries and MATLAB R2019a have been used for the implementation.}. The different steps involved in the pre-processing pipeline are depicted in Fig. \ref{fig:visualrepresentation} and explained below:

\begin{enumerate}[label=(\alph*)]
	\item {\it Face detection and tracking:} after extracting the frames, the Python library {\tt dlib} (v. 19.8.1) is used on the first video frame to obtain the ROI patch containing the face, as well as on every subsequent frame to detect the 68 facial landmarks. The three landmarks corresponding to the right eye lacrimal caruncle ($\boldsymbol{r}$), the left eye lacrimal caruncle ($\boldsymbol{l}$), and top nose ($\boldsymbol{n}$) are selected. A motion vector $\mathbf{\Delta}$ is then computed between each pair of consecutive frames by averaging the horizontal and vertical displacements of $\boldsymbol{r}$, $\boldsymbol{l}$ and $\boldsymbol{n}$, and smoothed temporally through a Savitzky-Golay filter on both dimensions \cite{Savitzky1964}. The initial patch is then tracked over time by shifting it of $\mathbf{\Delta}$ frame by frame.
	\item {\it Temporal partition:} after conversion to grayscale, overlapping temporal windows of $d$ seconds with a stride of $s$ seconds are isolated. This yields different temporal sequences of frames, whose numerosity depends on the duration of the video. A generic temporal sequence $\boldsymbol{S}$ resulting from this process is a 3D array of pixels of size $H\times W \times K$, where $H$ and $W$ depend on the output of the face detector on the first frame, and $K$ depends on the frame rate of the video. 
	\item {\it Area selection:} at this stage, we allow to select a specific  area of the face to be used for the feature analysis, in order to observe the relevance of different regions for the chosen feature representation. In our tests, we have considered three different cases, denoted in the following with upper-case letters (see Fig. \ref{fig:visualrepresentation}): the top-half ($T$), the bottom-half ($B$), or the full face information ($F$) is used.
\end{enumerate}

\subsection{Dynamic texture features}
\label{ssec:LDPTOP}

%
%

We aim at exploiting both spatial and temporal domains in the analysis of video sequences. To this purpose, we considered the Local Derivative Pattern features (LDP), already used for face recognition as a pattern descriptor  (e.g. \cite{Jabid2010,ZhangTIFS2010}), in their extended version involving the temporal domain, the Local Derivative Pattern on Three Orthogonal Planes (LDP-TOP) \cite{Phan2016FACESD}. 

The LDP, a generalization of the widely used Local Binary Pattern (LBP), is a point-wise operator applied to 2D arrays of pixels, that encodes diverse local spatial relationships. As suggested in \cite{ZhangTIFS2010}, we consider the second-order directional LDPs with direction $\alpha$, indicated as $\LDP^2_\alpha$, where $\alpha \in \{\ang{0},\, \ang{45},\, \ang{90},\, \ang{135} \}$.
Given a 2D array of pixels $A$, the $\LDP^2_\alpha$ at the location $(h,w)$ is an 8-bit vector defined as:

{\small
	\begin{align*}
		\label{eq:ldp}
		\LDP^2_\alpha(h,w) = [ f(I'_\alpha(h,w),I'_\alpha(h^-,w^-)),  f(I'_\alpha(h,w),I'_\alpha(h^-,w)), \\ \nonumber
		f(I'_\alpha(h,w),I'_\alpha(h^-,w^+)),  f(I'_\alpha(h,w),I'_\alpha(h,w^+)), \\  \nonumber
		f(I'_\alpha(h,w),I'_\alpha(h^+,w^+)),  f(I'_\alpha(h,w),I'_\alpha(h^+,w)), \\  \nonumber
		f(I'_\alpha(h,w),I'_\alpha(h^+,w^-)),  f(I'_\alpha(h,w),I'_\alpha(h,w^-)) ] \nonumber
	\end{align*}
}
with $h^+ \coloneqq h+1, h^- \coloneqq  h-1$ and $w^+ \coloneqq w+1, w^- \coloneqq w-1$. A representation of the $3\times 3$ neighborhood is depicted in Figure \ref{fig:ldp}(a).
The operator $I'_\alpha$ is the first-order derivative in the direction $\alpha$, and is defined pixel-wise as:
\begin{equation}
	I'_{\alpha}(h,w)= 
	\begin{cases}
		A(h,w) - A(h,w^+) & \text{if } \alpha=\ang{0} \\
		A(h,w) - A(h^-,w^+) &\text{if } \alpha=\ang{45} \\
		A(h,w) - A(h^-,w) & \text{if } \alpha=\ang{90}\\
		A(h,w) - A(h^-,w^-) & \text{if } \alpha=\ang{135}\\
	\end{cases}
\end{equation}
while
\begin{equation}
	f(a,b) = \begin{cases}
		0 & \text{if } x \cdot y > 0 \\
		1 & \text{if } x \cdot y \leq 0
	\end{cases}
\end{equation}

Essentially, $\LDP^2_\alpha(h,w)$ encodes whether first-order derivatives in the direction $\alpha$ have consistent signs when computed at $(h,w)$ and at proximal pixel locations. For a 2D array, the $\LDP^2_\alpha$ are extracted for every pixel and their $2^8$-bin histogram is computed; this is replicated for the four different directions, and the histograms are concatenated. 

Similarly as it is done in \cite{lbptop} for LBPs, in \cite{Phan2016FACESD} the authors propose to extend the computation of LDP histograms to 3D arrays. This is done by sequentially considering the three central 2D arrays along each dimension that intersect orthogonally (see Figure \ref{fig:ldp}(b)) and again concatenating the obtained histograms, yielding the so-called LDP-TOP features.

In our case, we apply this procedure to the temporal sequences $\boldsymbol{S}$ extracted as in Section \ref{ssec:prep}, and use the obtained histograms as features. Considering 4 derivative directions and three 2D arrays, the feature vector length is equal to $ 2^8 \times 4 \times 3 = 3072$.

In order to explore potential peculiarities in the way the temporal information is captured by LDPs, we add the opportunity to run the feature extraction on $\boldsymbol{S}$ in three different temporal modes, which differ by the orientation of the temporal information. In particular, we define:
\begin{itemize}
	\item {\it Direct mode} ($\rightarrow$): $\boldsymbol{S}$ is processed forward along the temporal direction; 
	\item {\it Inverse mode} ($\leftarrow$): $\boldsymbol{S}$ is processed backward along the temporal direction starting from the last frame;
	\item {\it Bidirectional mode} ($\leftrightarrow$): $\boldsymbol{S}$ is processed in both directions and histograms are concatenated (thus yielding a feature vector with doubled size).
\end{itemize}

\section{Classification  framework}
\label{sec:cl}

We now describe the framework adopted in our study for training a classifier and taking a decision on single tested videos.

\begin{figure*}[t!]
	\centering
	\begin{tikzpicture}[>=stealth]
	\node {\includegraphics[width=0.8\textwidth]{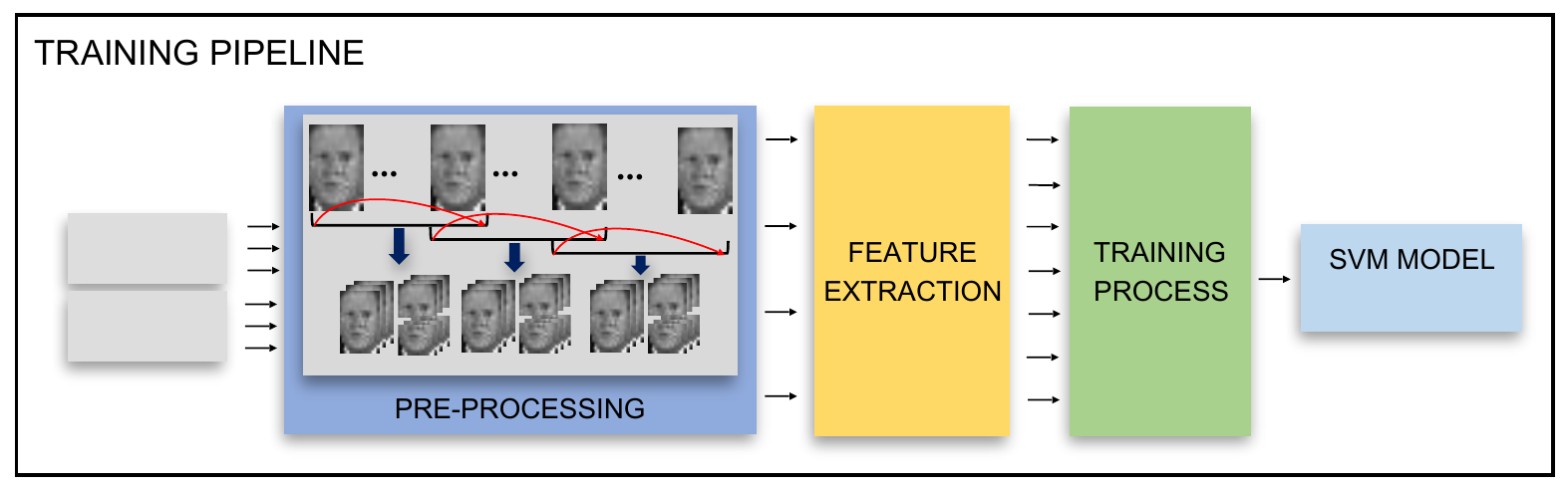}};	
	
	\node at (-5.95,-0.1) {$\tr_\text{r}$};
	\node at (-5.95,-0.8) {$\tr_\text{m}$};	
	
	\node at (5.9,-0.55) {$C$};		
	
	\end{tikzpicture}
	\caption{Training pipeline: given as input the training set of real and fake videos, provides as output the corresponding SVM model.}
	\label{fig:pipelinetraining}
\end{figure*}
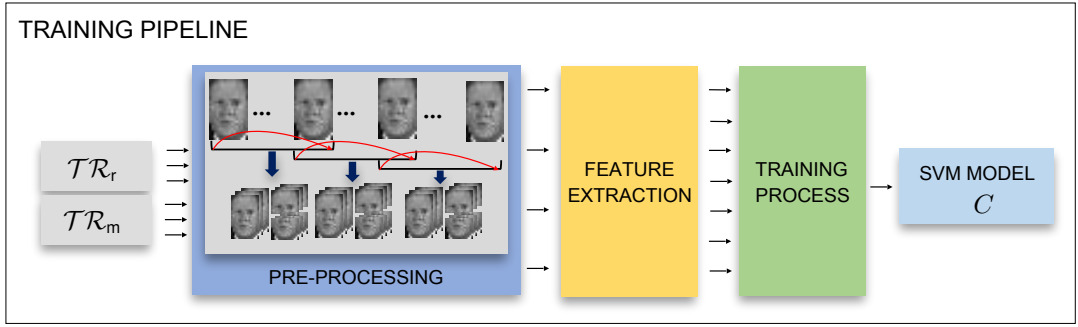

As depicted in Fig. \ref{fig:pipelinetraining}, the training process involves a set of real and manipulated videos, that we indicate as $\tr_\text{r}$ (labeled as $0$) and $\tr_\text{m}$ (labeled as $1$), respectively. Every video in these sets is fed into the pre-processing and the descriptor computation blocks, as described in Sections \ref{ssec:prep} and \ref{ssec:LDPTOP}. The feature vectors computed from each temporal sequences inherit the label of the video they belong and all of them are used as inputs for training the classifier $C$, a Support Vector Machines (SVM) with linear kernel\footnote{We used the MATLAB Statistics and Machine Learning Toolbox (v. R2019a) and selected a linear kernel function with predictor data standardization and Sequential Minimal Optimization (SMO).}.

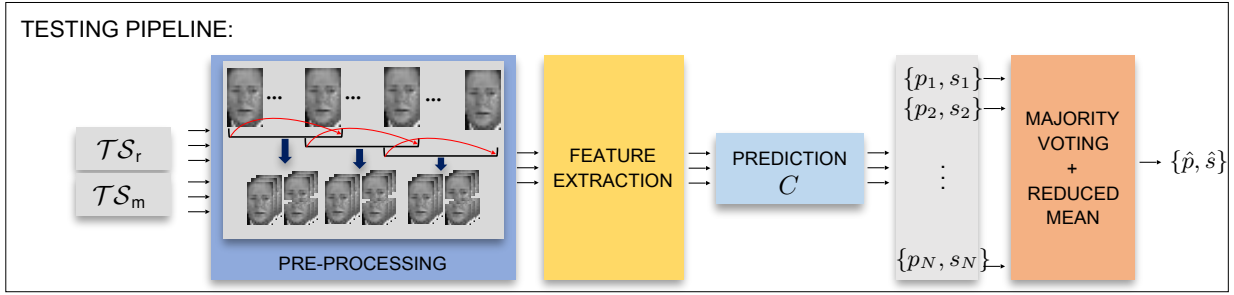
\begin{figure*}[t!]
	\centering
	\begin{tikzpicture}[>=stealth]
	\node at (0,0) {\includegraphics[width=0.9\textwidth]{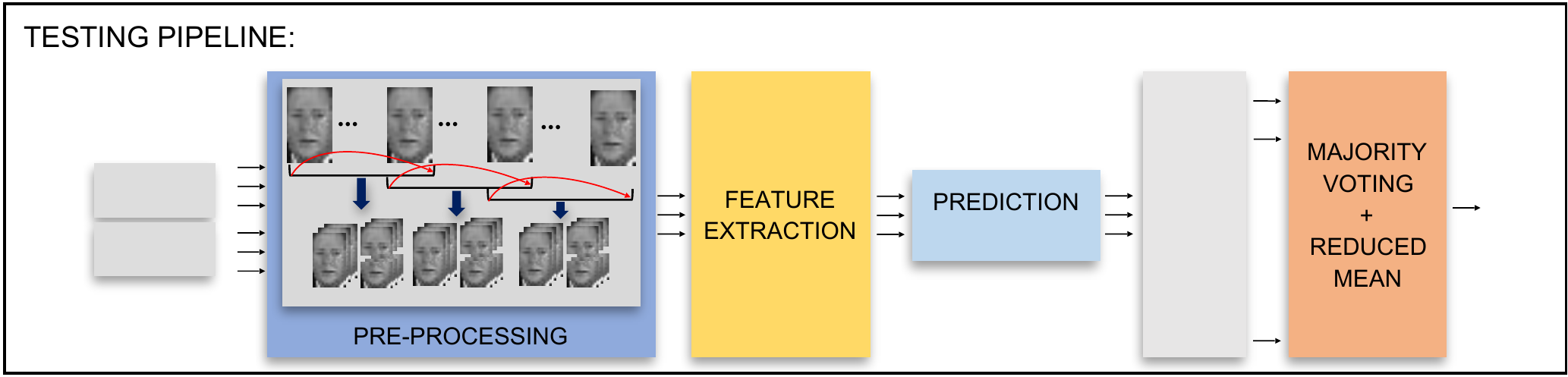}};
	
	\node at (-6.6,-0.05) {$\ts_\text{r}$};
	\node at (-6.6,-0.67) {$\ts_\text{m}$};	
	
	\node at (2.3,-0.5) {$C$};

	\node at (4.35,0.9) {\small$\{p_1,s_1\}$};		
	\node at (4.35,0.5) {\small$\{p_2,s_2\}$};			
	\node at (4.35,-0.3) {\small$\vdots$};	
	\node at (4.35,-1.5) {\small$\{p_N,s_N\}$};	
	
	\node at (7.75,-0.2) {\small$\{\hat{p},\hat{s}\}$};															
	
	\end{tikzpicture}
	\caption{Testing Pipeline: the pipeline $C$ returns a binary label $\hat{p}$ and the corresponding score $\hat{s}.$}
	\label{fig:testingpipeline}
\end{figure*}

Afterwards, the videos to be tested belong to sets that we will indicate as $\ts_\text{r}$ and $\ts_\text{m}$. The prediction on single videos is computed as depicted in Fig. \ref{fig:testingpipeline}. Pre-processing and descriptor computation are again performed and each resulting feature vector extracted is passed to the trained SVM model. 
This returns a pair $p_n,s_n$ for each of the $N$ temporal sequences extracted, where $p_n$ is the predicted label and $s_n$ is the output score of the SVM. In order to determine a final label $\hat{p}$ for the input video, a majority voting criterion is employed:
\begin{equation}
	\label{eq:maj}
	\hat{p}=\maj(\{p_1, \ldots, p_N\})
\end{equation}
where $\maj(\cdot)$ outputs the value recurring most frequently in the input set. In case of equal number of conflicting predictions, the $\maj$ criterion conservatively favors the $0$ class. 

Finally, for each video we compute a final score $\hat{s}$ through a ``reduced mean'' criterion: 
\begin{equation}
	\label{eq:rm}
	\hat{s} = \text{mean}(\{ s_n \text{ where } n \text{ is such that } p_n=\hat{p} \}),
\end{equation}
i.e., only the score values corresponding to the sequences whose predictions correspond to the final prediction $\hat{p}$ are averaged.


\section{Experimental results}
\label{sec:results}

The next sections present the experimental tests conducted in order to validate the proposed method in practical scenarios.

\begin{figure}[!b]
	\centering
	{\includegraphics[width=.45\linewidth]{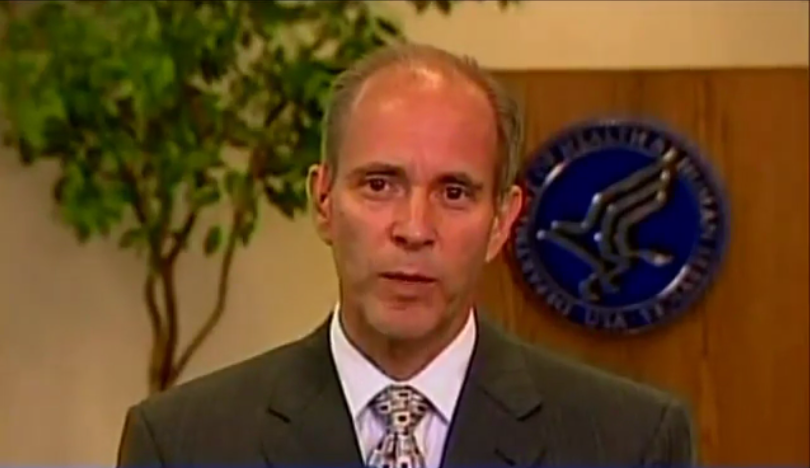}}\quad
	{\includegraphics[width=.45\linewidth]{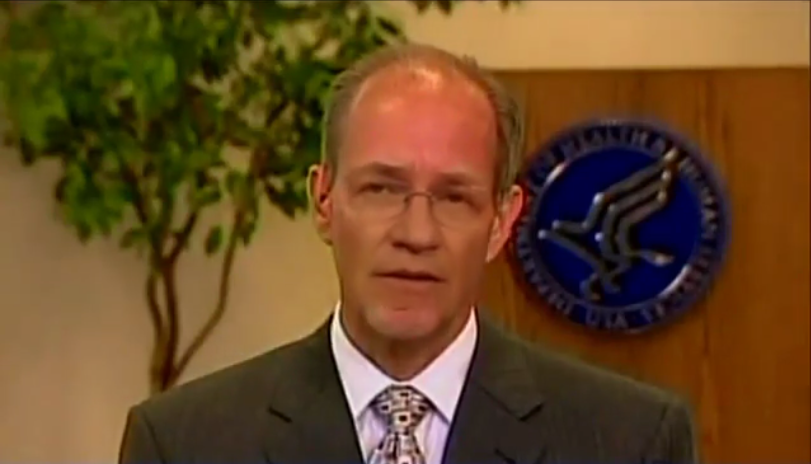}}\\
	{\includegraphics[width=.45\linewidth]{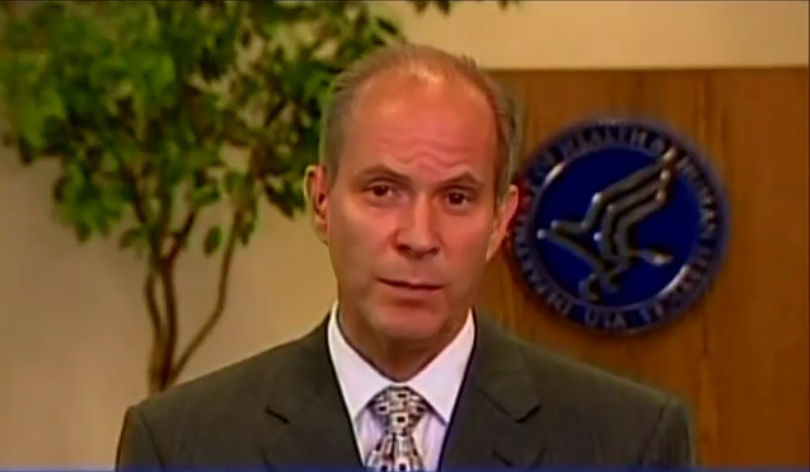}}\quad
	{\includegraphics[width=.45\linewidth]{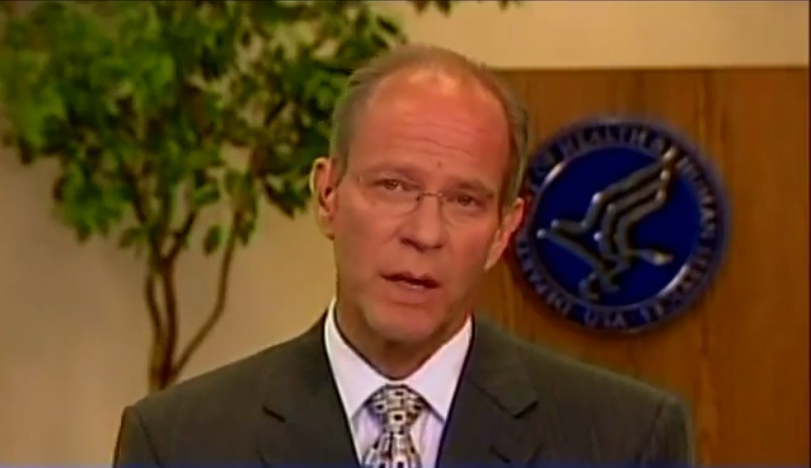}}
	\caption{{Frames extracted from (a) a sample $\OR$ video sequence and its (b) $\DF$, (c) $\FtF$ and (d) $\FSW$ manipulated versions.}}
	\label{fig:dataset}
\end{figure}

As a benchmark dataset of real and fake videos, we considered the FaceForensics++ dataset described in \cite{ff}, which consists of a large set of videos depicting human faces, which are then manipulated with different techniques. In particular, we have considered the 1000 original videos ($\OR$) and their manipulated counterparts through the \textit{Deepfake} ($\DF$) \cite{Deepfakeswebsite}, the \textit{Face2Face} ($\FtF$) \cite{Thies2019ACM} and the \textit{FaceSwap} ($\FSW$) \cite{Faceswapwebsite} techniques. We operate on the version of the dataset subject to a light compression (H.264 with constant rate quantization parameter equal to $23$). An example of these different manipulations is depicted in Fig. \ref{fig:dataset}. The videos are recorded under different conditions (e.g., interviews, TV shows, etc.), they have different length and are captured by different cameras. This results into a huge variability in terms of both data content and video structure (i.e., frame rate, video length, original coding standards, etc). 

The dataset comes with a standard split of videos for training, validation, and testing. In order to enable a fair comparison with other recently proposed approaches, we also considered the same training and testing set, yielding the sets $\tr_D$ with $|\tr_D|=720$ and $\ts_D$ with $|\ts_D|=140$, where $D \in \{ \OR, \DF, \FtF, \FSW \}$. Different subsets will be combined according to the experimental scenario considered. 

We have tested the feature representation and classification framework proposed in Section \ref{sec:method} and \ref{sec:cl} in several experimental scenarios and by analyzing different factors, which are described in details in the next subsections. For the sake of readability, we first summarize here the structure of our experimental validations:

\begin{itemize}
	\item {\bf Single-technique scenario} (Section \ref{ssec:1to1LDP23}). Original and fake videos are considered separately for different creation techniques; the impact of the temporal partition operation, the face area selection, and the temporal mode adopted are discussed.
	\item {\bf Multiple-technique scenario} (Section \ref{ssec:fusedoutcomeLDP23}). Videos created with arbitrary manipulation techniques are mixed in the testing; the capabilities of detecting and identifying the manipulation technique used in the testing phase is evaluated.
	\item {\bf Strong video compression} (Section \ref{ssec:comp}). The proposed detector is tested when a heavier compression is applied to the videos, thus its robustness against video compression is analyzed.
	\item {\bf Comparison with other descriptors} (Section \ref{ssec:lbp}). Performance comparison is discussed both considering the proposed detector exploiting the alternative spatio-temporal feature representation given by the LBP-TOP and other SoA approaches.
\end{itemize}


\subsection{Single-technique scenario}
\label{ssec:1to1LDP23}

\begin{figure*}[t!]
	\centering
	\includegraphics[width=0.8\textwidth]{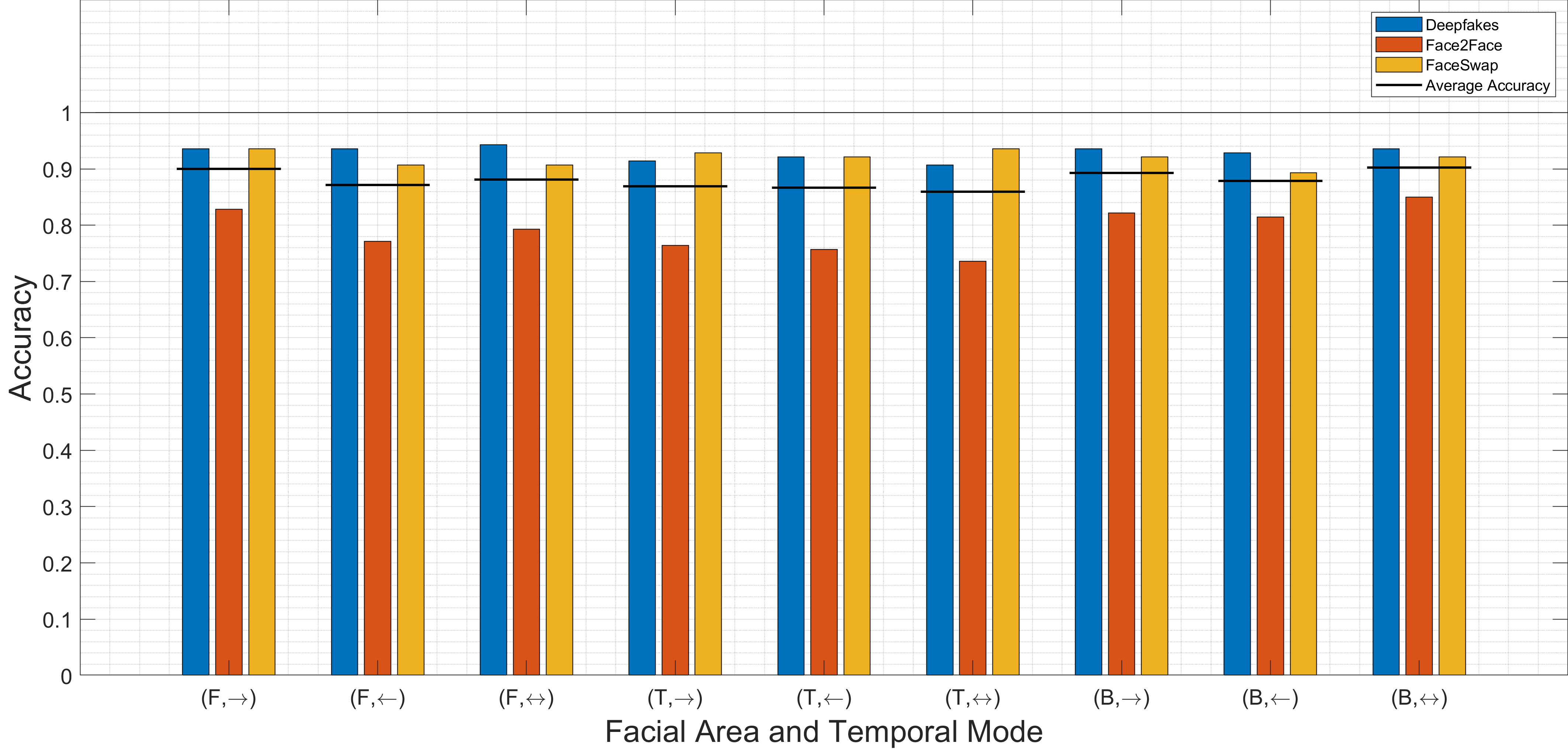}
	\caption{Classification accuracy per manipulation technique.}
	\label{fig:bar_LDP_23}
\end{figure*}

\begin{table*}[h!]
	\centering
	\caption{Classification accuracy and AUC computed on the single-manipulation scenario. Different facial areas and temporal modes are considered.}
	\label{tab:onetooneLDP23}
	\resizebox{0.8\linewidth}{!}{
		\begin{tabular}{c|c|c|c|c|c|c|c|c|}
			\cline{2-9}
			& \multicolumn{3}{c|}{\cellcolor[HTML]{ECF4FF}\textbf{Accuracy}} & \cellcolor[HTML]{ECF4FF}\textbf{\begin{tabular}[c]{@{}c@{}}Average\\ Accuracy\end{tabular}} & \multicolumn{3}{c|}{\cellcolor[HTML]{FFFFC7}\textbf{AUC}} & \cellcolor[HTML]{FFFFC7}\textbf{\begin{tabular}[c]{@{}c@{}}Average\\ AUC\end{tabular}} \\ \hline
			\multicolumn{1}{|c|}{\begin{tabular}[c]{@{}c@{}}Algorithm \\ Version\end{tabular}} & \textit{Deepfakes} & \textit{Face2Face} & \textit{FaceSwap} & \textbf{Cross-Dataset} & \textit{Deepfakes} & \textit{Face2Face} & \textit{FaceSwap} & \textbf{Cross-Dataset} \\ \hline
			\multicolumn{1}{|c|}{$(F,\dir)$} & 93,57\% & 82,86\% & 93,57\% & 90,00\% & 98,23\% & 88,08\% & 98,22\% &\textbf{ 94,85}\% \\ \hline
			\multicolumn{1}{|c|}{$(F,\inv)$} & 93,57\% & 77,14\% & 90,71\% & 87,14\% & 98,78\% & 85,14\% & 98,00\% & 93,97\% \\ \hline
			\multicolumn{1}{|c|}{$(F,\bid)$} & 94,29\% & 79,29\% & 90,71\% & 88,10\% & 98,65\% & 86,94\% & 97,45\% & 94,35\% \\ \hline
			\multicolumn{1}{|c|}{$(T,\dir)$} & 91,43\% & 76,43\% & 92,86\% & 86,90\% & 95,39\% & 78,13\% & 98,18\% & 90,57\% \\ \hline
			\multicolumn{1}{|c|}{$(T,\inv)$} & 92,14\% & 75,71\% & 92,14\% & 86,67\% & 94,41\% & 78,39\% & 98,00\% & 90,27\% \\ \hline
			\multicolumn{1}{|c|}{$(T,\bid)$} & 90,71\% & 73,57\% & 93,57\% & 85,95\% & 94,89\% & 80,78\% & 98,06\% & 91,24\% \\ \hline
			\multicolumn{1}{|c|}{$(B,\dir)$} & 93,57\% & 82,14\% & 92,14\% & 89,29\% & 97,53\% & 86,64\% & 97,47\% & 93,88\% \\ \hline
			\multicolumn{1}{|c|}{$(B,\inv)$} & 92,86\% & 81,43\% & 89,29\% & 87,86\% & 97,57\% & 85,91\% & 97,55\% & 93,68\% \\ \hline
			\multicolumn{1}{|c|}{$(B,\bid)$} & 93,57\% & 85,00\% & 92,14\% & \textbf{90,24}\% & 97,55\% & 88,63\% & 97,47\% & 94,55\% \\ \hline
		\end{tabular}
	}
\end{table*}

We tested the performance of our approach in separating original videos from videos that have been manipulated with a specific technique. The goal is to show the capabilities of each classifier when subjected to its corresponding test set. Thus:
\begin{align*}
	\tr_\text{r} = \tr_\OR & &  \tr_\text{m} = \tr_D \\
	\ts_\text{r} = \ts_\OR & &  \ts_\text{m} = \ts_D \\
\end{align*}
where $D$ varies in the set $\{\DF, \FtF, \FSW \}$. This yields an SVM classifier for every manipulation technique, that we denote as $C_{\DF}$, $C_{\FtF}$ and $C_{\FSW}$.

Videos in these sets are fed into the training pipeline described in Fig. \ref{fig:pipelinetraining}.
In this phase, we report the results obtained by employing the three different facial areas ($F$, $T$, and $B$) specified in Section \ref{ssec:prep} and the three temporal modes ($\dir$, $\inv$, and $\bid$) specified in Section \ref{ssec:LDPTOP}, yielding to nine classifiers per manipulation technique, to observe how they vary and interact.

Results are depicted as bar plots in Figure \ref{fig:bar_LDP_23} in terms of accuracy, i.e., the fraction of videos in $\ts_\text{r} \cup \ts_\text{m}$ that is assigned to the correct label. Full numerical results are reported in Table \ref{tab:onetooneLDP23}, where the value of the Area Under the Curve (AUC) obtained by thresholding $\hat{s}$ (i.e., the reduced-mean score) is also reported as performance indicator.

Tab. \ref{tab:onetooneLDP23} suggests that $C_{\DF}$ and $C_{\FSW}$ almost always allow for an accuracy greater that $90\%$, while for $C_{\FtF}$ the accuracy does not exceeds $85,0\%$. Interestingly, this correlates with the observations made in \cite{ff}, where a user study reveals that $\FtF$ generally produces more challenging manipulations to be detected for humans.

Moreover, it can be noticed that both the $F$ and the $B$ facial areas versions provide a better accuracy with respect to $T$. This indicates that the artifacts captured by the proposed feature representation are generally concentrated in the bottom part of the face. However, this effect is not uniform across manipulation techniques (see $\FSW$), suggesting that manipulation-specific patterns are likely introduced, as we will exploit in the next subsection.  

Finally, we observe that the inverse temporal mode alone does not introduce significant advantages, while the bidirectional mode generally does. This is not so suprising, given that the feature vector size is doubled, however the number of training samples remains the same.

In summary, the best results in terms of both performance indicators are achieved in the $(F,\dir)$ and the $(B,\bid)$ cases, respectively yielding $90,00\%$ and $90,24\%$ average accuracy. Therefore, for the sake of readability and space, we focus on the corresponding classifiers for the experimental analyses in the next subsections. 

As a further analysis, we evaluate the benefits of applying the temporal partition through sliding windows in the preprocessing phase by comparing with the baseline case where videos are not subdivided in shorter video sequences (i.e., the $w$ parameter in Fig. \ref{fig:visualrepresentation} is set equal to the video length in seconds) and only one LDP-TOP feature vector is extracted from each single video. This corresponds to the common approach of previously proposed detection methods (see \cite{ff}).

First, we observe in Table \ref{tab:numberofsamples} how the number of input feature vectors changes for these two cases, noticing that the sliding window approach increases the number of training/testing feature vectors by 6 to 8 times. Then, we provide in Table \ref{tab:slidingVSnoslidingLDP23} the accuracy loss when skipping the temporal partition step, defined as the difference in accuracy between of the ``sliding'' and ``non-sliding'' case (i.e., positive values indicate better performance of the ``sliding'' case). It can be noticed that the ``sliding'' approach always outperforms the ``non-sliding'' in terms of average accuracy among all datasets, with significant improvements (up to $10\%$) for $\FtF$. Just in some single cases, especially for $\FSW$, this observation is reversed, showing again manipulation-specific peculiarities. The two selected classifiers (top and bottom one in Table \ref{tab:slidingVSnoslidingLDP23}) however adhere to the general trend, showing an average accuracy increase of $3,33\%$ and of $2,86\%$.

\begin{table}[t!]
	\caption{Comparison between the number of samples (batches) obtained in case of non-sliding and sliding window approaches.}
	\label{tab:numberofsamples}
	\resizebox{\linewidth}{!}{
		\begin{tabular}{c|c|c|c|c|c|c|c|c|}
			\cline{2-9}
			& \multicolumn{2}{c|}{\OR} & \multicolumn{2}{c|}{\DF} & \multicolumn{2}{c|}{\FtF} & \multicolumn{2}{c|}{\FSW} \\ \cline{2-9} 
			& \textit{Training} & \textit{\begin{tabular}[c]{@{}c@{}}Testing\end{tabular}} & \textit{Training} & \textit{\begin{tabular}[c]{@{}c@{}}Testing\end{tabular}} & \textit{Training} & \textit{\begin{tabular}[c]{@{}c@{}}Testing\end{tabular}} & \textit{Training} & \textit{\begin{tabular}[c]{@{}c@{}}Testing\end{tabular}} \\ \hline
			\multicolumn{1}{|c|}{Sliding} & 3029 & 588 & 3026 & 588 & 2966 & 640 & 2307 & 482 \\ \hline
			\multicolumn{1}{|c|}{Non-sliding} & 360 & 70 & 360 & 70 & 360 & 70 & 360 & 70 \\ \hline
		\end{tabular}
	}
\end{table}

\begin{table}[t!]
	\small
	\centering
	\caption{Classification accuracy loss per manipulation technique when applying the "non-sliding" approach.}
	\label{tab:slidingVSnoslidingLDP23}
	\begin{tabular}{c|c|c|c|c|}
		\cline{2-5}
		& \multicolumn{3}{c|}{\cellcolor[HTML]{ECF4FF}\textbf{Accuracy Loss}} & \cellcolor[HTML]{ECF4FF}\textbf{\begin{tabular}[c]{@{}c@{}}Average\\ Accuracy Loss\end{tabular}} \\ \hline
		\multicolumn{1}{|c|}{\begin{tabular}[c]{@{}c@{}}Algorithm \\ Version\end{tabular}} & \DF & \FtF & \FSW & \textbf{Cross-Dataset} \\ \hline
		\multicolumn{1}{|c|}{$(F,\dir)$}& 1,43\% & 10,72\% & -2,14\% & 3,33\% \\ \hline
		\multicolumn{1}{|c|}{$(F,\inv)$}& 2,14\% & 5,71\% & -2,15\% & 1,90\% \\ \hline
		\multicolumn{1}{|c|}{$(F,\bid)$}& 2,15\% & 6,43\% & -4,29\% & 1,43\% \\ \hline
		\multicolumn{1}{|c|}{$(T,\dir)$}& 2,14\% & 10,00\% & 0,00\% & 4,04\% \\ \hline
		\multicolumn{1}{|c|}{$(T,\inv)$}& 2,14\% & 3,57\% & 0,00\% & 1,91\% \\ \hline
		\multicolumn{1}{|c|}{$(T,\bid)$} & 1,42\% & -1,43\% & 2,14\% & 0,71\% \\ \hline
		\multicolumn{1}{|c|}{$(B,\dir)$} & -0,72\% & 3,57\% & 0,00\% & 0,96\% \\ \hline
		\multicolumn{1}{|c|}{$(B,\inv)$}& 0,00\% & 5,72\% & -2,85\% & 0,96\% \\ \hline
		\multicolumn{1}{|c|}{$(B,\bid)$}& 1,43\% & 7,14\% & 0,00\% & 2,86\% \\ \hline
	\end{tabular}
	
\end{table}

\subsection{Multiple-technique scenario}
\label{ssec:fusedoutcomeLDP23}

We now consider the case where manipulation techniques are mixed. In particular, we approach the more realistic case where 
\begin{align*}
	\ts_\text{r} = \ts_\OR & &  \ts_\text{m} = \ts_{\DF} \cup \ts_\FtF \cup \ts_\FSW \\
\end{align*}
and the binary decision on each testing video needs to be taken blindly, i.e., without prior information on the manipulation technique used.

\begin{figure*}[t!]
	\centering
	\begin{tikzpicture}[>=stealth]
	\node at (0,0) {\includegraphics[width=0.9\textwidth]{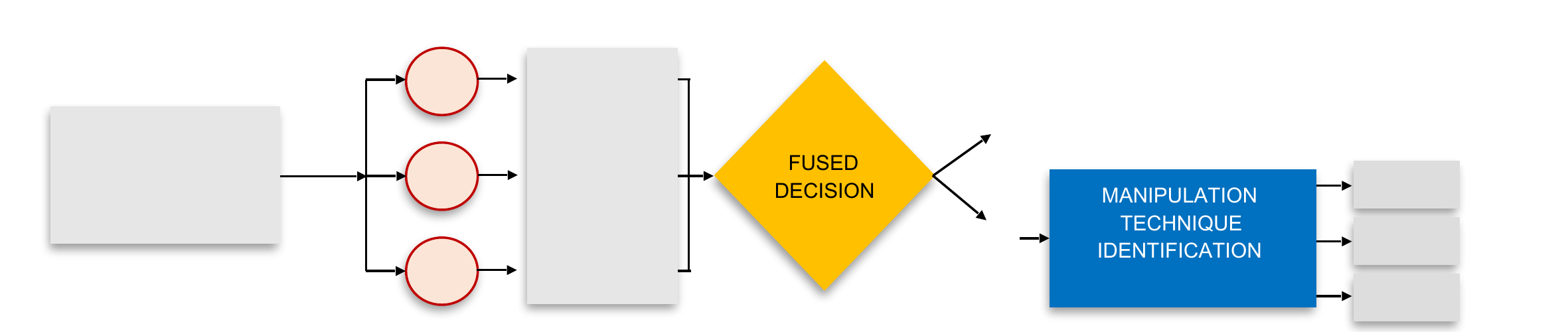}};
	
	\node at (-7.1,0.3) {$\ts_\OR$};
	\node at (-5.9,0.3) {$\ts_\DF$};	
	\node at (-7.1,-0.5) {$\ts_\OR$};
	\node at (-5.9,-0.5) {$\ts_\DF$};		
	
	\node at (-3.6,0.9) {\small$C_\DF$};
	\node at (-3.6,-0.1) {\small$C_\FtF$};	
	\node at (-3.6,-1.1) {\small$C_\FSW$};		
	
	\node at (-1.9,0.9) {\small$\{\hat{p}_\DF,\hat{s}_\DF\}$};			
	\node at (-1.9,-0.1) {\small$\{\hat{p}_\FtF,\hat{s}_\FtF\}$};			
	\node at (-1.9,-1.1) {\small$\{\hat{p}_\FSW,\hat{s}_\FSW\}$};	
	
	\node at (2.35,0.4) {\small 0};							
	\node at (2.35,-0.7) {\small 1};		
	
	\node[white] at (4.25,-1.25) {\small$\max\{ \hat{s}_\DF, \hat{s}_\FtF,  \hat{s}_\FSW\}$};
	
	\node at (6.55,-0.2) {\small \DF};	
	\node at (6.55,-0.8) {\small \FtF};
	\node at (6.55,-1.4) {\small \FSW};																																			
	
	\end{tikzpicture}
	\caption{Decision pipeline for the multiple-technique scenario.}
	\label{fig:FusedDecision}
\end{figure*}
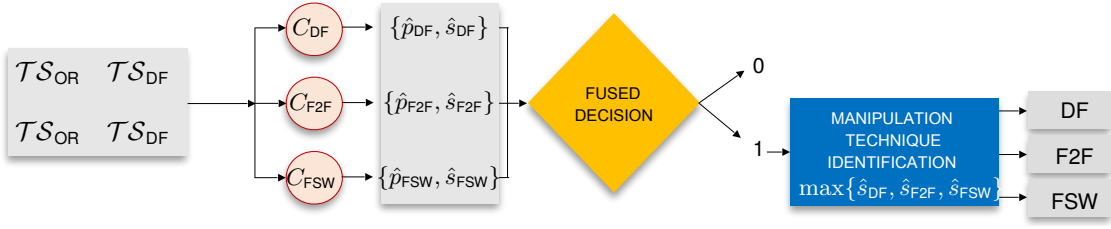

We have experienced that training a single binary classifier with $\tr_\text{r} = \tr_\OR$ and $\tr_\text{m} = \tr_{\DF} \cup \tr_\FtF \cup \tr_\FSW$ brings to poor results. This might be interpreted in view of the linearity of the classifier used, which seemingly does not allow to properly separate the two classes through an hyperplane in the feature space. Instead of enforcing that a single classifier can accurately separate the samples, we rather propose to combine the outcome of classifiers trained on single manipulation techniques. This also allows us to estimate the used manipulation technique in case of positive detection in a cascade fashion as represented in Figure \ref{fig:FusedDecision}.

More specifically, we propose to assign each test video a label $\hat{p} \in \{0,1\}$ by combining the outputs of the classifiers $C_{\DF}$, $C_{\FtF}$ and $C_{\FSW}$ trained as in Section \ref{ssec:1to1LDP23}. This yields to three predicted labels $\hat{p}_{\DF}$, $\hat{p}_{\FtF}$, $\hat{p}_{\FSW}$, and three average scores $\hat{s}_{\DF}$, $\hat{s}_{\FtF}$, $\hat{s}_{\FSW}$. 
Then, the three estimated labels are passed to a fusion block that applies the logical OR operator (indicated as $\vee$) in order to get $\hat{p}$. In other words, a video is classified as manipulated as soon as one of the three detectors returns the label $1$. Furthermore, in case of $\hat{p}=1$, the maximum value of the scores is selected as indicator of the manipulation technique used to create the video. 

Table \ref{tab:overallresults3SVM23LDP} reports the accuracy results obtained through this approach for the two variants selected in Section \ref{ssec:1to1LDP23}, $(F,\dir)$ and $(B,\bid)$, which consistently exceed $85\%$. We also report the false positive rate (fraction of original videos erroneously classified as manipulated) and the false negative rate (fraction of manipulated videos erroneously classified as original). The former seems to be more crucial for this fused approach, likely due to the fact that original videos are underrepresented in the overall training set.

%

\begin{table}[b!]
	\small
	\centering
	\caption{Classification results in the multiple-technique scenario.}
	\label{tab:overallresults3SVM23LDP}
	\begin{tabular}{|c|c|c|c|}
		\hline
		\textbf{\begin{tabular}[c]{@{}c@{}}Algorithm\\ Version\end{tabular}} & 
		\textbf{\begin{tabular}[c]{@{}c@{}}False \\Positive\\ Rate\end{tabular}} & \textbf{\begin{tabular}[c]{@{}c@{}}False \\Negative\\ Rate\end{tabular}} & \textbf{Accuracy} \\ \hline
		$(F,\dir)$  & 20,00\% & \textbf{11,43\%} & 86,43\% \\ \hline
		$(B,\bid)$  & 15,71\% & 11,90\% & \textbf{87,14\%} \\ \hline
	\end{tabular}
\end{table}


Finally, we measure the accuracy in estimating the manipulation technique used when a video is correctly classified as manipulated. Table \ref{tab:confusion23LDPFULLDIRECT} and Table \ref{tab:confusion23LDPBOTTOMDIRECTINVERSE} are the confusion matrices of the two classifiers for this task. 
The high diagonal values (around $90,00\%$ in most cases) indicate that the feature representation carries quite strong information on the specific manipulations techniques. 

\begin{table}[b!]
	\small
	\centering
	\caption{Confusion matrix for the manipulation estimation task with $(F,\dir)$.}
	\label{tab:confusion23LDPFULLDIRECT}
	\begin{tabular}{cc|c|c|c|}
		\cline{3-5}
		&  & \multicolumn{3}{c|}{\cellcolor[HTML]{FFFFFF}\textbf{\scriptsize PREDICTIONS}} \\ \cline{3-5} 
		&  & \cellcolor[HTML]{FFFFFF}{\DF} & \cellcolor[HTML]{FFFFFF} {\FtF} & \cellcolor[HTML]{FFFFFF} {\FSW} \\ \hline
		\multicolumn{1}{c}{\cellcolor[HTML]{FFFFFF}} & \cellcolor[HTML]{FFFFFF} {\DF} & \cellcolor[HTML]{94DF9D}\textbf{89,23} & \cellcolor[HTML]{F17373}10,77\%& \cellcolor[HTML]{F17373} 0,00\% \\ \cline{2-5} 
		\multicolumn{1}{c}{\cellcolor[HTML]{FFFFFF}} & \cellcolor[HTML]{FFFFFF} {\FtF} & \cellcolor[HTML]{F17373} 5,17\% & \cellcolor[HTML]{94DF9D}\textbf{91,38\%} & \cellcolor[HTML]{F17373} 3,45\% \\ \cline{2-5} 
		\multicolumn{1}{c}{\multirow{-3}{*}{\cellcolor[HTML]{FFFFFF}\rotatebox[origin=c]{90}{\textbf{\scriptsize TARGET}}}} & \cellcolor[HTML]{FFFFFF} {\FSW} & \cellcolor[HTML]{F17373} 0,00\% & \cellcolor[HTML]{F17373}3,17\% & \cellcolor[HTML]{94DF9D}\textbf{96,83\%} \\ \hline
	\end{tabular}
\end{table}

\begin{table}[b!]
	\small
	\centering
	\caption{Confusion matrix for the manipulation estimation task with $(B,\bid)$.}
	\label{tab:confusion23LDPBOTTOMDIRECTINVERSE}
	\begin{tabular}{cc|c|c|c|}
		\cline{3-5}
		&  & \multicolumn{3}{c|}{\cellcolor[HTML]{FFFFFF}\textbf{\scriptsize PREDICTIONS}} \\ \cline{3-5} 
		&  & \cellcolor[HTML]{FFFFFF}{\DF} & \cellcolor[HTML]{FFFFFF} {\FtF} & \cellcolor[HTML]{FFFFFF} {\FSW} \\ \hline
		\multicolumn{1}{c}{\cellcolor[HTML]{FFFFFF}} & \cellcolor[HTML]{FFFFFF} {\DF} & \cellcolor[HTML]{94DF9D}\textbf{83,33} & \cellcolor[HTML]{F17373}16,17\%& \cellcolor[HTML]{F17373} 0,00\% \\ \cline{2-5} 
		\multicolumn{1}{c}{\cellcolor[HTML]{FFFFFF}} & \cellcolor[HTML]{FFFFFF} {\FtF} & \cellcolor[HTML]{F17373} 5,08\% & \cellcolor[HTML]{94DF9D}\textbf{91,53\%} & \cellcolor[HTML]{F17373} 3,39\% \\ \cline{2-5} 
		\multicolumn{1}{c}{\multirow{-3}{*}{\cellcolor[HTML]{FFFFFF}\rotatebox[origin=c]{90}{\textbf{\scriptsize TARGET}}}} & \cellcolor[HTML]{FFFFFF} {\FSW} & \cellcolor[HTML]{F17373} 0,00\% & \cellcolor[HTML]{F17373}5,00\% & \cellcolor[HTML]{94DF9D}\textbf{95,00\%} \\ \hline
	\end{tabular}
\end{table}

\subsection{Impact of Strong Video Compression}
\label{ssec:comp}

\begin{figure*}[t!]
	\centering
	\includegraphics[width=0.8\textwidth]{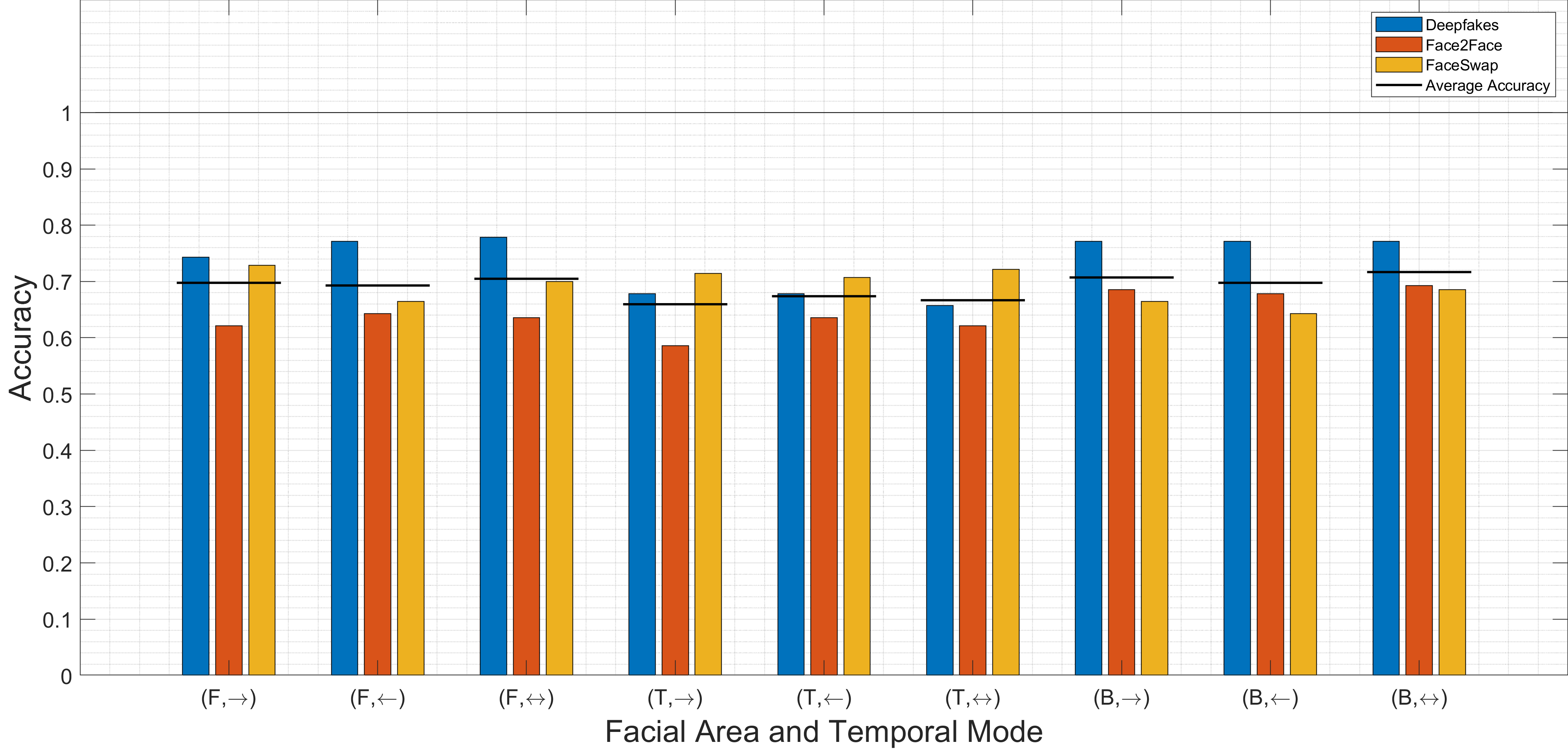}
	\caption{Classification accuracy per manipulation technique in case of strong video compression.}
	\label{fig:bar_LDP_40}
\end{figure*}

\begin{table*}[t!]
	\centering
	\caption{Classification accuracy and AUC computed on the single-manipulation scenario in case of strong video compression.}
	\resizebox{0.8\linewidth}{!}{
		\begin{tabular}{c|c|c|c|c|c|c|c|c|}
			\cline{2-9}
			& \multicolumn{3}{c|}{\cellcolor[HTML]{ECF4FF}\textbf{Accuracy}} & \cellcolor[HTML]{ECF4FF}\textbf{\begin{tabular}[c]{@{}c@{}}Average\\ Accuracy\end{tabular}} & \multicolumn{3}{c|}{\cellcolor[HTML]{FFFFC7}\textbf{AUC}} & \cellcolor[HTML]{FFFFC7}\textbf{\begin{tabular}[c]{@{}c@{}}Average\\ AUC\end{tabular}} \\ \hline
			\multicolumn{1}{|c|}{\begin{tabular}[c]{@{}c@{}}Algorithm \\ Version\end{tabular}} & \textit{Deepfakes} & \textit{Face2Face} & \textit{FaceSwap} & \textbf{Cross-Dataset} & \textit{Deepfakes} & \textit{Face2Face} & \textit{FaceSwap} & \textbf{Cross-Dataset} \\ \hline
			\multicolumn{1}{|c|}{$(F,\dir)$} & 74,29\% & 62,14\% & 72,86\% & 69,76\% & 80,49\% & 68,97\% & 80,59\% & 76,68\% \\ \hline
			\multicolumn{1}{|c|}{$(B,\bid)$} & 77,14\% & 69,29\% & 68,57\% & \textbf{71,67}\% & 81,35\% & 74,04\% & 79,64\% & \textbf{78,34}\% \\ \hline
		\end{tabular}
	}
	\label{tab:onetooneLDP40}
\end{table*}

The FaceForensics++ dataset also offers a more heavily compressed version of the videos, i.e., with $cf = 40$. As reported in \cite{ff}, the quality degradation due to compression compromises the performance of detection algorithms, as well as humans. We therefore assess how this impacts our method by reproducing the single-technique scenario for the two best performing classifiers, and report the results in Figure \ref{fig:bar_LDP_40} and Table \ref{tab:onetooneLDP40}. While keeping an average accuracy around $70\%$, the performance decrease is evident when compared to Figure \ref{fig:bar_LDP_23} (around $20\%$), thus confirming that, as most of the existing methods, our feature representation also suffers from the application of a heavier compression. This holds also for the multiple-technique scenario, where the accuracy of the best classifier drop to $71\%$ (see Table \ref{tab:overallresults3SVM40LDP}).

\begin{table}[t!]
	\small
	\centering
	\caption{Classification accuracy in the multiple-technique scenario in case of strong video compression.}
	\label{tab:overallresults3SVM40LDP}
	\begin{tabular}{|c|c|c|c|c|}
		\hline
		\textbf{\begin{tabular}[c]{@{}c@{}}Algorithm\\ Version\end{tabular}} & \textbf{\begin{tabular}[c]{@{}c@{}}Decision\\ Criterion\end{tabular}} & \textbf{\begin{tabular}[c]{@{}c@{}}False \\ Positive\\ Rate\end{tabular}} & \textbf{\begin{tabular}[c]{@{}c@{}}False \\ Negative\\ Rate\end{tabular}} & \textbf{Accuracy} \\ \hline
		$(F,\dir)$ & $\vee$ & 50,00\% & 27,62\% & 66,79\% \\ \hline
		$(B,\bid)$ & $\vee$ & 42,86\% & \textbf{23,81\%} & \textbf{71,43\%} \\ \hline
	\end{tabular}
\end{table}

\subsection{Comparison with other descriptors}
\label{ssec:lbp}

\begin{figure*}[t!]
	\centering
	\includegraphics[width=0.8\textwidth]{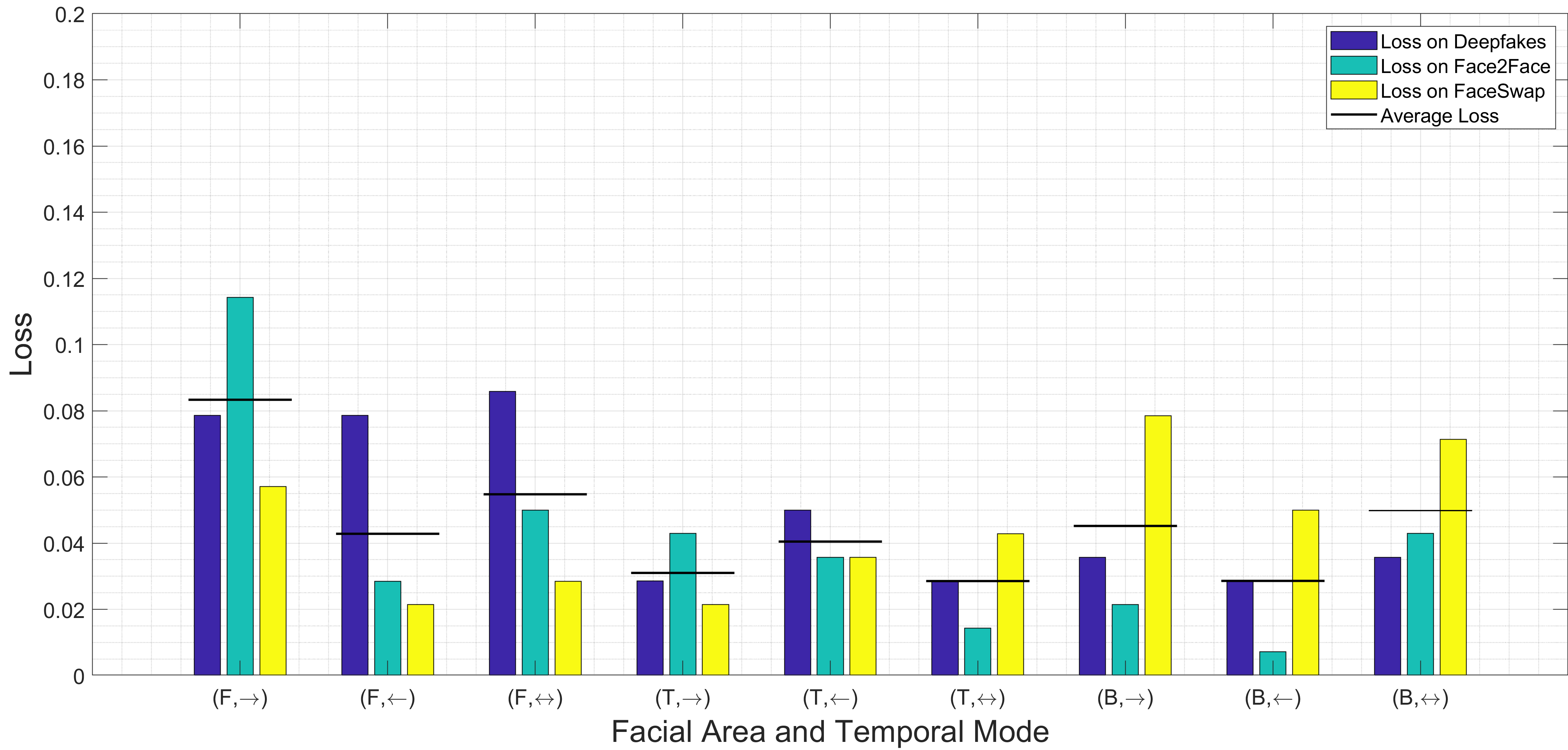}
	\caption{Classification accuracy loss per manipulation technique when using LBP-TOP descriptors instead of the proposed ones.}
	\label{fig:loss_LBP_23}
\end{figure*}

\begin{figure*}[t!]
	\centering
	\begin{tikzpicture}
	\node {\includegraphics[width=\textwidth]{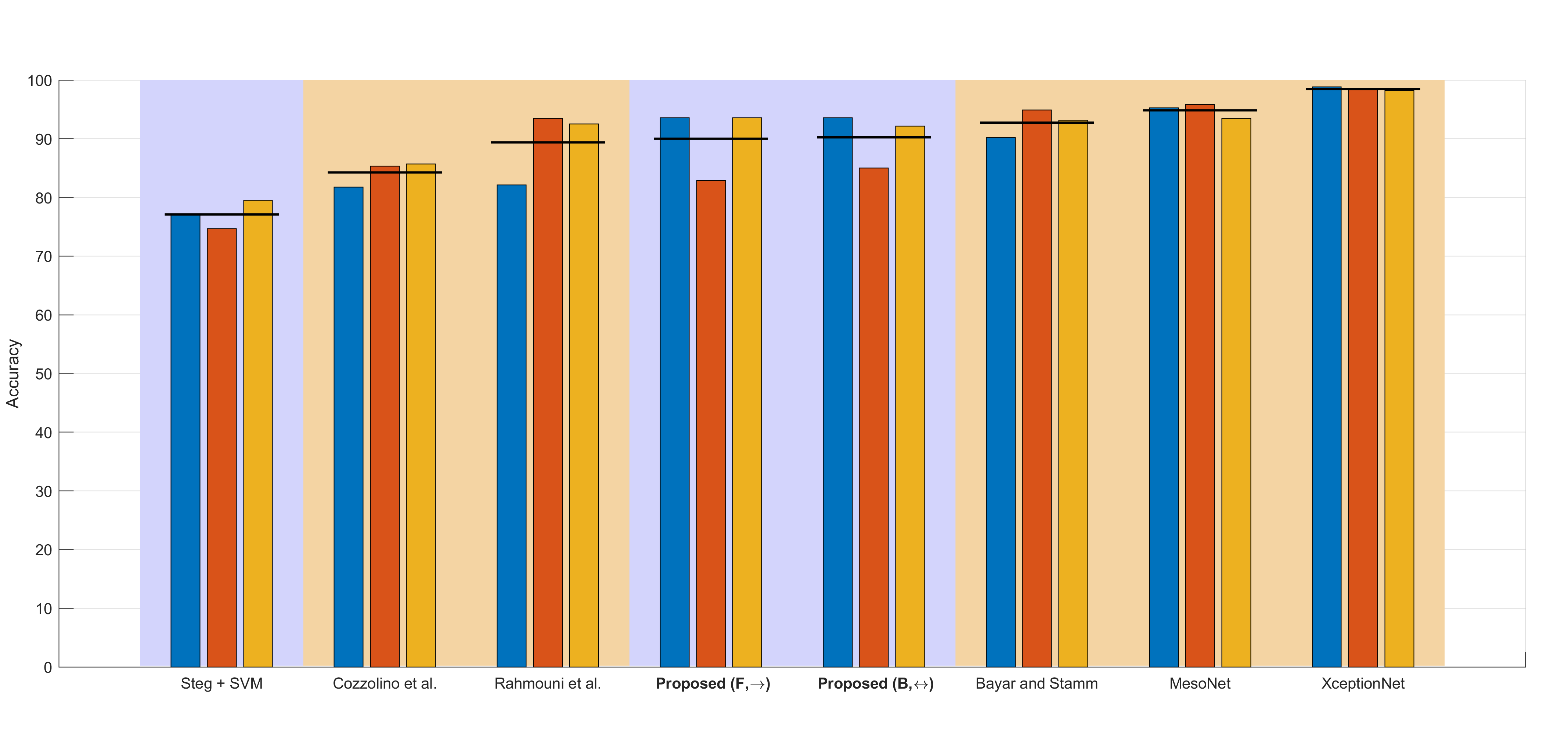}};
	
	\node at (-6,2.9) {\includegraphics[width=0.1\textwidth]{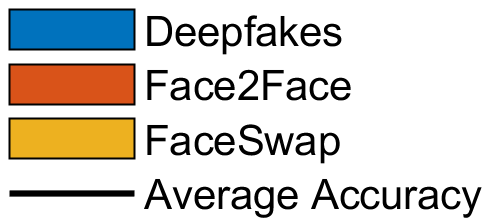}};
	
	\end{tikzpicture}
	
	\caption{Classification accuracies of the proposed algorithms with respect to other detection methods. Orange background indicates that the method is based on CNNs.}
	\label{fig:bar_comparison}	
\end{figure*}

In this subsection, we consider the performance of our method with respect to other detection algorithms.

We first compare our feature representation with a known competitor among the spatio-temporal texture descriptors used in face anti-spoofing, the LBP-TOP \cite{Zhao2007}. Differently from LDPs, LBPs capture only information on the first-order directional derivatives computed at a central reference pixel, that are thresholded, encoded into a binary number, and finally collected into histogram over different pixels; LBP-TOP is the corresponding temporal extension and yields a feature vector of length $[1,177]$, obtained by applying the uniform pattern version of the LBP features that led to a more compact feature vector and descriptor robust to rotations. 
We want to determine whether and how much the improved performance observed in \cite{Phan2016FACESD} for the face spoofing detection task generalizes to the detection of facial manipulations. To this purpose, the tests performed in Section \ref{ssec:1to1LDP23} are extended by replacing the LDP-TOP feature vector with the LBP-TOP one, while keeping unchanged all the other steps described in Sections \ref{sec:method} and \ref{sec:cl}. 
We report in Fig. \ref{fig:loss_LBP_23} the classification accuracy loss observed when using LBP-TOP instead of LDP-TOP (i.e., with respect to the results in Figure \ref{fig:bar_LDP_23}). The loss is always positive, thus LDP-TOP indeed outperforms LBP-TOP by a significant margin. 


Then, we position our results with respect to other methods proposed in literature and benchmarked on the same dataset in \cite{ff}. Since the training, validation, and testing splits of the FaceForensics++ dataset are standard, it is fair to compare the results obtained through our proposed pipelines with the ones reported in \cite{ff} in terms of accuracy on the testing set. Figure \ref{fig:bar_comparison} reports the results of our $(F,\rightarrow)$ and $(B,\leftrightarrow)$ classifiers and other six detection methods, namely ``Steg+SVM'' \cite{steg}, ``Cozzolino et al.'' \cite{cozzolino}, ``Rahmouni et al.'' \cite{Rahmouni2017}, ``Bayar and Stamm'' \cite{bayar}, ``MesoNet'' \cite{Afchar2018}, and ``XceptionNet'' \cite{ff}, sorted according to their average accuracy over manipulation techniques. All of them, except for the ``Steg+SVM'', are based on convolutional neural networks.
Remarkably, our approach outperforms the SVM-based one \cite{steg} by a large margin, and also two techniques based on CNNs \cite{cozzolino} and \cite{Rahmouni2017}. While the performance of other deep networks like XceptionNet remains significantly better, the proposed spatio-temporal descriptors, separated linearly in the feature space, provide fairly accurate results with the advantages of higher explainability of the encoded patterns and limited training time.

\section{Conclusions}
\label{sec:conc}
In this paper we have proposed a novel methodology to detect fake video sequences by exploiting spatio-temporal descriptors successfully exploited for the task of face anti-spoofing. Results show good performance on various manipulation techniques and in different experimental scenarios. Relatively small feature representation and relatively simple classifiers allow to detect manipulated video sequences and identify the adopted manipulation technique. 

Future work will deal with the challenging problem of heavy video compression, where current literature still does not achieve satisfactory results. Moreover, further extension will consider the scenario where new manipulation techniques could be considered and learned by the detector, e.g. by exploiting innovative paradigms coming from the machine learning domain like incremental learning.

\section*{Acknowledgments}

This work was supported by the project PREMIER (PREserving Media trustworthiness in the artificial Intelligence ERa), funded by the Italian Ministry of Education, University, and Research (MIUR) within the PRIN 2017 program. The second author was partially supported by Archimedes Privatstiftung, Innsbruck.

\bibliographystyle{abbrv}
\bibliography{references}
\end{document}